\documentclass[10pt,twocolumn,letterpaper]{article}

\usepackage{wacv}
\usepackage{times}
\usepackage{epsfig}
\usepackage{graphicx}
\usepackage{amsmath}
\usepackage{amssymb}
\usepackage{multirow}
\usepackage{amsthm}
\usepackage{caption}
\usepackage{subcaption}
\usepackage{pifont}
\usepackage[dvipsnames]{xcolor}

\definecolor{mygray}{gray}{0.6}
\newcommand{\xmark}{\ding{51}}%
\def\wacvPaperID{1062} % Enter the WACV Paper ID here

\wacvfinalcopy % *** Uncomment this line for the final submission
\ifwacvfinal
\def\assignedStartPage{9876} % *** Enter the assigned starting page number (instead of 9876)
\fi

\ifwacvfinal
\usepackage[breaklinks=true,bookmarks=false]{hyperref}
\else
\usepackage[pagebackref=true,breaklinks=true,colorlinks,bookmarks=false]{hyperref}
\fi

\ifwacvfinal
\else
\fi

\begin{document}

\newtheorem*{suppt}{Supplementary role of context}
\newtheorem*{relativet}{Relative role of context}

%%%%%%%%% TITLE
\title{Towards Contextual Learning in Few-shot Object Classification }

\author{Mathieu Pagé Fortin \\
Laval University, QC, CA\\
{\tt\small mathieu.page-fortin.1@ulaval.ca}
\and Brahim Chaib-draa \\
Laval University, QC, CA\\
{\tt\small brahim.chaib-draa@ift.ulaval.ca}
}

\maketitle
%\thispagestyle{empty}

%%%%%%%%% ABSTRACT
\begin{abstract}
    Few-shot Learning (FSL) aims to classify new concepts from a small number of examples. While there have been an increasing amount of work on few-shot object classification in the last few years, most current approaches are limited to images with only one centered object. On the opposite, humans are able to leverage prior knowledge to quickly learn new concepts, such as semantic relations with contextual elements. 
    
Inspired by the concept of contextual learning in educational sciences, we propose to make a step towards adopting this principle in FSL by studying the contribution that context can have in object classification in a low-data regime. To this end, we first propose an approach to perform FSL on images of complex scenes. We develop two plug-and-play modules that can be incorporated into existing FSL methods to enable them to leverage \textit{contextual learning}. More specifically, these modules are trained to weight the most important context elements while learning a particular concept, and then use this knowledge to ground visual class representations in context semantics. Extensive experiments on Visual Genome and Open Images show the superiority of \textit{contextual learning} over learning individual objects in isolation.
\end{abstract}

%%%%%%%%% BODY TEXT
\section{Introduction}

Whereas Convolutional Neural Networks are currently the state-of-the-art models for object recognition tasks, they generally require a large number of examples from each class to perform well. In the last few years an increasing amount of effort has been done in zero-shot learning (ZSL) and few-shot learning (FSL) to develop approaches that reduce the number of examples required to train efficient models. Progress in this direction is important for solving many real-world problems for which labeling is hard, and for enabling new applications, such as robots that actively learn new concepts on the fly from their environment \cite{part2017incremental}.

However, most current FSL methods focus on visual features and tend to only consider objects in isolation~\cite{al2016recovering,chen2019image,li2019large,snell2017prototypical,sung2018learning, vinyals2016matching,wang2019tafe,xing2019adaptive,zhang2019few}. These methods are therefore primarily evaluated on datasets of images with only one centered object (e.g. \textit{mini}Imagenet~\cite{vinyals2016matching}, Omniglot~\cite{lake2015human}, CUB-200 \cite{welinder2010caltech}). On the other hand, images in real-world applications can be more complex, containing many different objects. This scenario has been neglected so far, and while it can be more relevant for real applications, we argue that complex scenes also offer an interesting opportunity to supplement visual information with contextual and semantic relations between concepts. This idea is motivated by the principle of \textit{contextual learning} \cite{johnson2002contextual} in educational sciences, and more specifically by the first functional feature of context defined by Dohn \etal. \cite{dohn2018concept}: 

\begin{suppt} $\left[Context\right]$ is brought in, or added to, the understanding of a phenomenon—the focal object—that would not have been adequately understood had it been considered in isolation. A context thus completes the conditions for understanding the focal object. \cite{dohn2018concept}
\end{suppt}

In the few-shot setting, the model has access to a set of \textit{base} classes with many examples and is evaluated on its ability to learn \textit{novel} classes from few examples. This means that when a \textit{novel} class is presented in its context to the model, some \textit{base} classes can also appear in the scene. This is similar to when humans see an unknown object in a familiar scene, for instance a corkscrew in a kitchen. To learn this new label, our brain will not only process the appearance of the object, but also the contextual and semantic relations with other objects \cite{bar2004visual}, for instance a bottle of wine. This is fundamental to learn quickly and for continual incremental learning in humans, as new concepts are generally not learned in isolation, but often in relation to already known concepts \cite{bar2004visual, oliva2007role}. We apply this principle to FSL by proposing a method to add contextual semantic information in visual representations: our model learns to refine class prototypes according to the context in which training examples appear. 

However, this low-data setting poses the challenge of context generalization: \textit{How to model the context such that this can generalize to \textit{novel} classes with only few examples of scenes?} Intuitively, one could represent the context of a class $i$ by co-occurrence counts, i.e. by counting how many times the class $i$ occurs in the same scene as all other classes in the training set. However, this is restrictive for at least two reasons. First, co-occurrence counts assume that the training set is a sufficient sample to estimate the large variety of co-occurrences that can happen, which is certainly not the case in a few-shot setting. Second, co-occurrence counts are limited to statistical regularities and they ignore the potential semantic relations within each class, which could help building powerful representations.

Instead, we make use of transfer learning from a text model pretrained on large corpora to represent the semantics of classes. Previous work has shown that word embeddings implicitly encode high-level relations between entities~\cite{gupta2017distributed, mikolov2013distributed, zablocki2019context}, which could help context generalization in a few-shot setting by enabling the model to capture higher-order relations. For instance, if in the small training set plates co-occur with forks, a semantic-aware model could infer that plates are also likely to co-occur with spoons because of the semantic similarity between forks and spoons. On the other hand, if there are irrelevant entities in the scene, such as ``wall'', which is generally not very informative, or out-of-context objects that could induce confusion \cite{rosenfeld2018elephant}, an ideal context-aware model should ignore them. These aspects of \textit{contextual learning} are defined by the second functional feature of context of Dohn \etal. \cite{dohn2018concept}:

\begin{relativet}
The context is centered around the object. The context is not a neutral layout of things or
properties near the focal object, nor is it a set of circumstances or an indefinite “background”. It is
ordered and organized by its relations to the focal object, which co-determines what properties of the
surroundings are relevant and thus part of the context. \cite{dohn2018concept}
\end{relativet}

We integrate this principle in our approach by proposing a Class-conditioned Context Attention Module (CCAM) such that our model can learn to attend to context elements that are relevant to the focal object. Psychological studies showed that contextual cueing in humans improves object classification in scenes, by capitalizing on the fact that most objects co-occur more often with certain objects and not others \cite{oliva2007role}. But additionally, the relative role of context specifies that not all co-occurrences are equal \cite{dohn2018concept}. Our experiments show that CCAM effectively weights discriminative contextual objects more strongly.

In summary, our main contributions are the following:
\begin{itemize}
    \item We propose a few-shot model that learns class representations grounded in contextual semantics. To this end, we propose a gated visuo-semantic unit (GVSU), a flexible module to combine visual prototypes with contextual semantic information. 
    \item We propose CCAM, an attention module applied on context elements that automatically learns to attend to the most important entities in scenes relatively to the focal object. 
    \item We conduct extensive experiments on Visual Genome~\cite{krishnavisualgenome} and Open Images \cite{OpenImages}, which are large-scale datasets of complex scenes with hundreds of classes. Our results support that using context is valuable in a few-shot setting. 
    \item As an auxiliary result, we observe that our model implicitly learns semantic word embeddings grounded in scene context. 
\end{itemize}

\section{Related Work}
FSL approaches can be divided into \textit{gradient-based} methods \cite{finn2017maml, finn2018probabilistic, jamal2019taml} and \textit{metric learning based} methods \cite{Bateni_2020_CVPR, gidaris2018dynamic,li2019revisiting, snell2017prototypical, sung2018learning}. Gradient-based methods aim to improve the training procedure, which MAML \cite{finn2017maml} is a typical example. MAML is a meta-learning algorithm that aims to generalize such that new tasks can be learned with few update steps. On the opposite, metric learning approaches aim to learn a metric space where support (train) examples are embedded such that query (test) examples can be classified based on a distance metric (e.g. euclidean distance \cite{snell2017prototypical}, cosine similarity \cite{gidaris2018dynamic}, or Mahalanobis distance~\cite{Bateni_2020_CVPR}) without requiring any parameter update to learn novel classes. Our work is more closely related to metric learning and that is why we focus on this family of approaches for the rest of this section. 

\paragraph{Few-shot image classification.}
Several FSL approaches build on Prototypical Networks \cite{snell2017prototypical}. This method learns a metric space by computing class centroids from the examples in the support set. It then compares query image embeddings with these prototypes and assigns a class by performing nearest neighbor search. Other approaches consider different ways to compare support and query embeddings, such as Relation Networks \cite{sung2018learning} that automatically learn the distance function with a neural network, or the approach of Li \textit{et al.} \cite{li2019revisiting} that compares support and query images based on several descriptors.

Other approaches leverage relations between classes to transfer visual features of \textit{base} classes to \textit{novel} ones. For instance, Wang \textit{et al}. \cite{wang2018zero} proposed to use a Graph Convolutional Network to transfer features between classes based on a knowledge base that encodes relations between these categories. Li \textit{et al}. \cite{li2019large} developed an approach that learns from predicting class hierarchies, which facilitates feature transfer. A similar idea has been proposed to transfer explicit attributes between classes~\cite{al2016recovering}. In our work, we use another form of transfer learning between classes. We leverage the presence of \textit{base} classes when \textit{novel} classes are presented in a scene to adapt their representation in metric space. By doing so, our model can benefit from contextual cueing \cite{oliva2007role} when a new instance is seen in a complex scene.

\paragraph{Auxiliary semantics in FSL.} Recently, additional cues that were only considered in ZSL have proved to be also useful in FSL, especially when the quantity of examples for each class is very low. Xing \textit{et al.} \cite{xing2019adaptive} built on Prototypical Networks~\cite{snell2017prototypical} by adding word embeddings in the formation of class prototypes in their approach called AM3. This improves the accuracy in $1$-shot by almost $10\%$ on \textit{mini}Imagenet \cite{vinyals2016matching} and by $5\%$ on CUB-200 \cite{welinder2010caltech}. Furthermore, Schwartz \textit{et al}.~\cite{schwartz2019baby} built on AM3 by additionally using text descriptions of classes extracted from WordNet, and thereby improved $1$-shot accuracy by an additional $2\%$ on \textit{mini}Imagenet. 

\paragraph{Context semantics.} All the FSL work cited above focus on visual information. Indeed, even the attributes, word embeddings and text descriptions that are used in these work need to encode features that can be detected visually from the appearance of a new concept. However, this is not the only form of semantic information that can be available in images. The presence of other objects in a scene can also inform which classes are more or less likely to appear~\cite{bar2004visual, oliva2007role,zablocki2019context}. The context has been used recently to improve object detection within deep learning models in a standard setting with large datasets \cite{chen2017spatial, liu2018structure, mottaghi2014role,woo2018linknet}. 

Recently Zablocki \textit{et al.} \cite{zablocki2019context} introduced the use of scene context in ZSL. They showed that their model, with the use of Word2vec \cite{mikolov2013distributed} embeddings, could learn to rank unseen classes according to their likelihood of appearing in an image given the presence of other objects. This suggests that word embeddings implicitly encode co-occurrences of other classes in real visual scenes, even if they have been trained on text corpora \cite{mikolov2013distributed}. This is closely related to the distributional hypothesis \cite{harris1954distributional}, which states that \textit{words that appear in similar contexts often have similar meanings}. This is exploited by skip-gram and continuous bag-of-words (CBOW) models, and the results from Zablocki \textit{et al.} \cite{zablocki2019context} suggest that this principle could also generalizes to visual scenes: \textit{items denoted by words that have similar meanings tend to occur in similar scenes}. This idea has been explored by Lüddecke \textit{et al.} \cite{luddecke2019distributional}, where they proposed a method to learn semantic word embeddings explicitly from images showing objects in context. 

To the best of our knowledge, scene context has not been used in previous FSL work. In this paper, we argue that FSL would benefit from considering objects with their context, because it places new concepts in relation with previously acquired knowledge. FSL models could then leverage these relations to benefit from contextual cueing \cite{oliva2007role} in new scenes. We introduce this idea of using scene context in FSL to perform object classification in complex images by building on Prototypical Networks~\cite{snell2017prototypical} and by learning class prototypes grounded in context. Unlike Zablocki \textit{et al.} \cite{zablocki2019context}, our model jointly learns to embed visual information with context semantics and word embeddings of class labels. Furthermore, whereas some recent semantic-based approaches use relations between classes to share common visual features, our use of scene context with word embeddings exploit a different form of semantic relation which is complementary and orthogonal to visual features. 

\section{Our model}

\begin{figure*}[t]
    \centering
    \includegraphics[width=\linewidth]{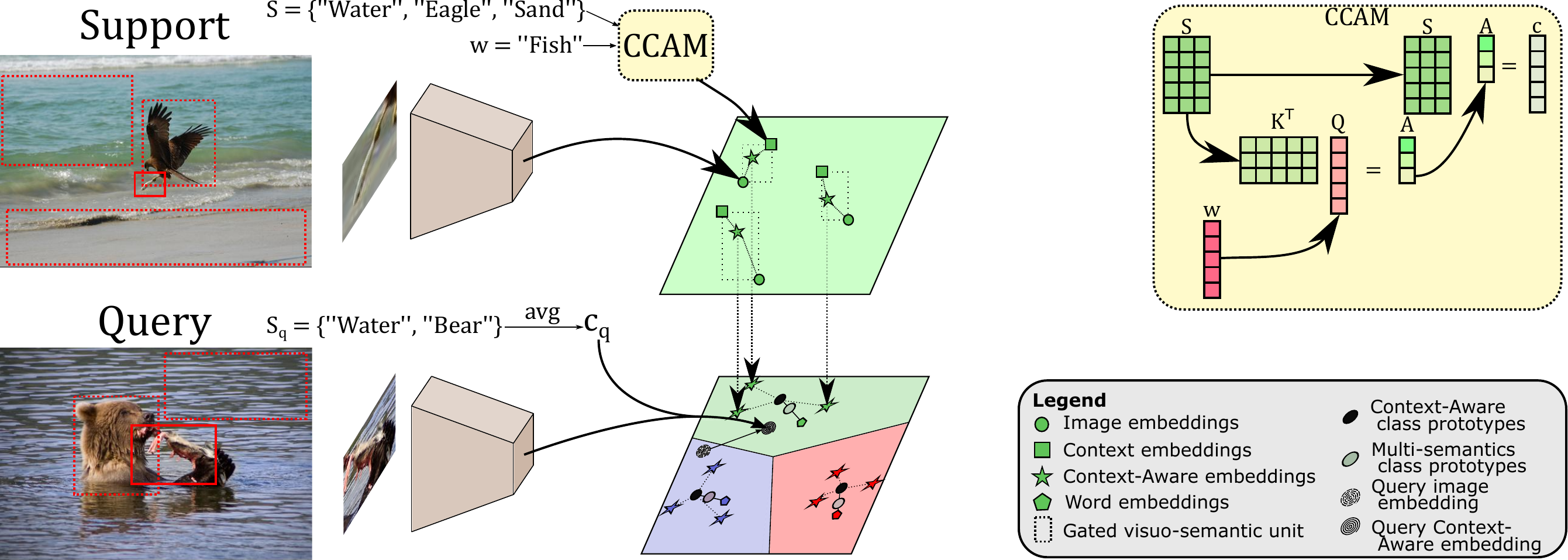}
    \caption{\small Overview of our model. The focal object (red box with solid lines) is cropped and embedded by a CNN. CCAM computes the relative role of support context elements according to the word embedding $w$ of the class label. Image and context embeddings are projected in a same space (represented by a circle and a square, respectively), and our GVSU produces context-aware embeddings (represented by a star). All context-aware embeddings for a given class are averaged to produce context-aware class prototypes (black dots). Finally multi-semantics class prototypes (light dots) are obtained by refining them with their word embeddings.}
    \label{fig:overview}
\end{figure*}
\subsection{Preliminaries}
FSL aims to solve the problem of $M$-way $K$-shot classification, where $M$ is the number of classes in a given task and $K$ is a small number of examples for each class. Generally, few-shot models are trained on a large dataset $\mathcal{D}_{train}$ with a set of \textit{base} classes that is disjoint from the \textit{novel} categories in $\mathcal{D}_{test}$. The goal is to learn a representation model $f_\theta$ on $\mathcal{D}_{train}$ such that it can learn to recognize novel categories with only $K$ examples. This is generally done by simulating the episodic test scenario of $M$-way $K$-shot classification during training. That is, even if a large number of examples are available for each class at train time, $f_\theta$ is trained by sampling at each episode $e$ (1) a \textit{support} set $\mathcal{S}_e = \{(x_i, y_i)\}_{i=1}^{M\times K}$ that contains $K$ examples for each $M$ class and (2) a \textit{query} set $\mathcal{Q}_e = \{(q_j, y_j)\}_{j=1}^{n_q}$ containing $n_q$ images of the same classes sampled in the support set. The model is then trained according to the cross-entropy loss:
\begin{equation}
    \mathcal{L}(\theta) = - \frac{1}{n_q} \sum_{t=1}^{n_q}\log p_\theta(y_t|q_t, \mathcal{S}_e)
\end{equation}

Prototypical networks \cite{snell2017prototypical} offer a simple and efficient way to model $p_\theta(y|q, \mathcal{S}_e)$. Each of the images in the support set are embedded by a CNN denoted by $f_\theta : \mathbb{R}^D \rightarrow \mathbb{R}^{d_x}$. Then a prototype is built for each class by averaging the $K$ vector embeddings from the same class: 
\begin{equation}
    \textbf{c}_k = \frac{1}{|\mathcal{S}_k|}\sum_{(x_i,y_i)\in \mathcal{S}_k} f_\theta(x_i)    
\end{equation}

Finally, the class distribution of a query image $q$ is assigned by computing the softmax over the euclidean distances $d$ of its embedding $f_\theta(q)$ and all class prototypes $\textbf{c}_k$:
\begin{equation}
    p_\theta(y=k|q, \mathcal{S}_e) = \frac{\exp(-d(f_\theta(q), \textbf{c}_k))}{\sum_{k'}\exp(-d(f_\theta(q), \textbf{c}_{k'}))}    
\end{equation}

\subsection{Context-Aware prototypes learning}
Following previous work on the supplementary role of context in learning \cite{dohn2018concept} and classification \cite{bar2004visual,oliva2007role}, we propose to learn class prototypes that embed knowledge about their context. To achieve that, we augment the support and query sets with scene context $S$, $\mathcal{S}_e = \{(x_i, S_i,y_i)\}_{i=1}^{M\times K}$, $\mathcal{Q}_e = \{(q_j, S_j,y_j)\}_{j=1}^{n_q}$, and we adapt the formulation of each prototype $\textbf{c}_k$ as:
\begin{equation}
    \hat{\textbf{c}}_k = \frac{1}{|\mathcal{S}_k|}\sum_{(x_i,y_i)\in\mathcal{S}_k}\phi(f(x_i), g(c_i)),
    \label{eq:cawareproto}
\end{equation}

where $c_i$ is the context representation of object $i$ obtained from CCAM (see below for more details), $g(c_i)$ is a small neural network projecting $c_i$ in the same space than the image embedding, and $\phi(\cdot,\cdot)$ is a function that adapts the image embedding according to the scene context (see section~\ref{gated}). An overview of our approach is shown in Fig.~\ref{fig:overview}. We now describe these components in more details.

\subsection{Class-conditioned Context Attention Module (CCAM)}
\paragraph{Scene context.}
 We model the scene context by converting surrounding objects from \textit{base} classes into semantic vector representations. This is done by leveraging word embeddings learned from a semantic model such as Word2vec \cite{mikolov2013distributed} pre-trained on Wikipedia. Therefore, the scene context of an object is represented by the matrix $S \in \mathbb{R}^{d_w \times n_s}$, where $d_w$ is the word embeddings dimension and $n_s$ is the number of surrounding objects. 

\paragraph{Class-conditioned Context Attention.}
The relative role of context \cite{dohn2018concept} suggests that some elements are more important than others when understanding a particular object. For instance, the concept of \textit{bathroom} might be important with respect to a \textit{toilet}, but could dupe the model while learning the concept of \textit{cat's paw}, even if in the few support examples cats are observed in bathrooms (e.g. see Fig. \ref{ccamc}). To respect this phenomenon, we propose a Class-conditioned Context Attention Module (CCAM) that enables our model to weight the importance of each context elements in $S$ while learning a particular concept $w$ (see CCAM in Fig.~\ref{fig:overview}). This is done by computing a scaled dot-product attention score~$A$~\cite{vaswani2017attention} between the word embedding $w$ of the class label and each element in $S$ after linear transformations:
\allowdisplaybreaks
%\begin{equation}
\begin{align*}
    K &= W_KS \\
    Q & = W_Qw \tag{\stepcounter{equation}\theequation} \\
    A & = softmax(\frac{K^\intercal Q}{\sqrt{d_c}}) \\
    c & = SA ,
    \end{align*}
%\end{equation}
where $W_K, W_Q \in \mathbb{R}^{d_c\times d_w}$ are weights matrices, and $\frac{1}{\sqrt{d_c}}$ is a scaling factor proposed in \cite{vaswani2017attention} to obtain smoother scores. $A$ reflects the relative role of each object in $S$, which is used to weight the contribution of context entities with respect to the focal object.

\paragraph{Context averaging $C_{avg}$.} Note that the attention mechanism in CCAM is exclusively applied on context from the support set since it depends on the class category $w$, which is unknown in queries. For query instances, the context representation $c_q$ is simply obtained by averaging all class embeddings $w_q$ in $S_q$: 

\begin{equation}
    c_q = \frac{1}{n_s}\sum_{w_q \in S_q} w_q 
    \label{eq:cavg}
\end{equation} 

\subsection{Gated visuo-semantic unit (GVSU)}
\label{gated}
To combine visual embeddings with context semantics according to the supplementary role of context \cite{dohn2018concept}, we propose a gated visuo-semantic unit, a module to adaptively combine each feature individually from both representations based on a gating mechanism. This is modeled by:

\begin{equation}
\begin{split}
    \phi(f(x), g(c)) & := z \cdot f(x) + (1 - z) \cdot g(c)\\
    h_v & = \tanh(W_v \cdot f(x))\\
    h_c & = \tanh(W_c \cdot g(c))\\
    z & = \sigma(W_z \cdot [h_v, h_c]),\\
    \end{split}
\end{equation}
where $W_v \in \mathbb{R}^{d_z \times d_x}, W_c \in \mathbb{R}^{d_z \times d_c}, W_z \in \mathbb{R}^{d_x \times 2 d_z}$ are weights matrices, and $\sigma$ is the sigmoid function. 

Our fusion mechanism is in the same vein than the Gated Multimodal Unit (GMU) \cite{arevalo2017gated} that proved to be successful with multimodal representations. Our module slightly differs in that the output representation of the original GMU is $h~=~z\cdot~h_v~+~(1-z)~\cdot~h_c$, whereas in our formulation the intermediate representations $h_v$ and $h_c$ are used to compute the weighting factors to apply on each dimension of $f(x)$ and $g(c)$. The goal of our GVSU is to move the image embeddings in the semantic space according to the context representation, such that objects of the same class will cluster together based on their appearance \textit{and} their contextual semantics. 

\subsection{Multi-semantics prototypes}
Lastly, we add the word embedding of the class label to further refine each prototype. We adopt a similar mechanism that Xing \textit{et al}. \cite{xing2019adaptive} proposed to combine visual prototypes with their word embeddings, since it proved to be particularly useful in settings with less data. Our context-aware prototypes are thus refined by:

\begin{equation}
    \textbf{c}_k' = \lambda\cdot\hat{\textbf{c}}_k + (1 - \lambda)\cdot \hat{w}_k
\end{equation}
where $\hat{w}_k$ is a transformation of the word embedding $w_k$ and $\lambda$ is a coefficient between 0 and 1. Both $\hat{w}_k$ and $\lambda$ are obtained with a two-layer neural network that uses $w_k$ as input.

Finally, the class distribution of a query image $q$ is computed as:
\begin{equation}
    p(y=k|q, S_q, \mathcal{S}_e, w) \propto  \exp(-d[\phi(f(q), g(c_q)), \textbf{c}_k'])
\end{equation}
In summary, the intuition of our approach is to model the prototypes $\textbf{c}_k'$ according to the appearance of support examples, the context in which they appear, \textit{and} the semantics of the class. Moreover, the context of support examples is represented as a weighted sum of the word embeddings of the \textit{base} classes that co-occur in the scene. To perform inference, a query $q$ is compared to these prototypes $\textbf{c}_k'$ based on their euclidean distance. The appearance of the query object is fused with the context of the scene, which is modeled as the average of the word embeddings of \textit{base} classes in the scene. 

\subsection{Assumptions}
Note that our approach assumes that the context is known, similar to \cite{zablocki2019context} did in a zero-shot setting, as the detection of objects is another task upstream to the FSL problem on which our work focuses. One could remove this assumption by replacing ground-truth annotations with an object detection module. Indeed, note that we form the context with \textit{base} classes only, for which training examples can be in large number during the metric-learning phase. Our approach could also be complementary to the growing number of work on few-shot object detection \cite{Jang_2020_WACV, Li_2020_CVPR, Perez_Rua_2020_CVPR}, where classification is generally performed for each object individually without considering the context.

\begin{table*}[]
\caption{\small Average Top-1 accuracy (\%) with $95\%$ confidence intervals on Visual Genome and Open Images . Results are averaged over 4000 test episodes for Visual Genome and 1000 test episodes for Open Images. V: Uses visual information; W: Uses word embeddings; C: Uses contextual information.}
    \centering
        \begin{tabular}{c|l|ccc|cccc}
         \hline
         Dataset&\textbf{Model} &V&W&C& $5$-way $1$-shot & $5$-way $5$-shot & $20$-way $1$-shot & $20$-way $5$-shot\\
         \hline
        \multirow{3}{*}{Visual Genome} & PN \cite{snell2017prototypical} &\xmark&&& $52.23 \pm 0.76$&$69.37 \pm 0.63$&$25.71 \pm 0.29$&$42.61 \pm 0.70$\\
        & AM3 \cite{xing2019adaptive}&\xmark&\xmark&&$62.50 \pm 0.66$&$72.07 \pm 0.74$&$34.36 \pm 0.32$&$ 44.84 \pm 0.61$\\
         \cline{2-9}
        & Ours&\xmark&\xmark&\xmark&$\textbf{71.54} \pm \textbf{0.57}$&$ \textbf{78.50} \pm \textbf{0.55}$&$ \textbf{46.13} \pm \textbf{0.47}$&$ \textbf{54.72} \pm \textbf{0.49}$\\
         \hline
         \hline
       \multirow{3}{*}{Open Images}&  PN \cite{snell2017prototypical}&\xmark&& &$64.60 \pm 1.43$& $80.44 \pm 1.25 $& $34.93 \pm 0.98$&$52.73 \pm 1.02$\\
        & AM3 \cite{xing2019adaptive}&\xmark&\xmark&& $69.87 \pm 1.00$&$80.43 \pm 1.02$& $40.25 \pm 0.47$& $52.59 \pm 0.94$\\
         \cline{2-9}
        & Ours&\xmark&\xmark&\xmark&$ \textbf{77.42} \pm \textbf{1.22}$   &$\textbf{87.70} \pm \textbf{0.95}$&$\textbf{51.61} \pm \textbf{0.99}$& $\textbf{67.53} \pm \textbf{0.88}$\\
         \hline
    \end{tabular}
    \label{fig:openimages_res}
\end{table*}

\section{Experiments}
\subsection{Dataset and settings}
 Traditional FSL datasets such as \textit{mini}Imagenet \cite{vinyals2016matching} and CUB-200 \cite{welinder2010caltech} mainly contain images with only one object and little context. Therefore, we rather experiment on Visual Genome \cite{krishnavisualgenome} and Open Images \cite{OpenImages}, which are large datasets of scenes with several objects in each image.
\paragraph{Visual Genome \cite{krishnavisualgenome}.}
We randomly split the images in $70\%/10\%/20\%$ \textit{train}, \textit{validation} and \textit{test} sets, respectively. We start by using the public splits by Zablocki \etal. \cite{zablocki2019context} that keep $50\%$ of classes for \textit{base+val} classes and $50\%$ for \textit{novel} classes. However, a closer look at those sets showed that some \textit{novel} classes are very similar to \textit{base} classes, which could bias generalization evaluations. For instance, ``bottle'' and ``television'' are in the \textit{base} set, but ``bottles'' and ``TV'' are \textit{novel} classes. To solve this issue, we filter the \textit{novel} classes whose Word2vec \cite{mikolov2013distributed} embeddings have a cosine similarity higher than $0.75$ with any of the \textit{base} or \textit{val} classes. It effectively removes singular/plural nouns and closely related concepts such as ``police'' (\textit{base} class) and ``policeman'' (\textit{novel} class). This will prevent our model from picking on those biases that would overestimate FSL performance.

We use the bounding box annotations to crop image parts that correspond to objects, and we remove examples whose smallest side is less than $25$ pixels. Following this, we remove the classes that appear in less than $10$ images. We also form a set of validation classes for hyper-parameter search. This finally results in $969$ \textit{base} classes, $242$ \textit{val} classes and $829$ \textit{novel} classe.

\paragraph{Open Images v6 \cite{OpenImages}.} We start by using the original \textit{train}/\textit{val}/\textit{test} splits\footnote{https://storage.googleapis.com/openimages/web/download.html} of images. Then, we randomly sample $400/100/100$ \textit{base}/\textit{val}/\textit{novel} classes, respectively. Classes that appear less than 10 times in their respective split of images are removed and the \textit{novel} classes are also filtered based on the Word2vec \cite{mikolov2013distributed} cosine similarity with \textit{base} and \textit{val} classes (see above). This results in $371/38/57$ \textit{base}/\textit{val}/\textit{novel} classes. 

\subsection{Implementation details}
We employ a ResNet-12 CNN backbone as described in~\cite{oreshkin2018tadam} and we train it from scratch. It is made of $4$ blocks with $3$ layers of $3\times3$ convolutions and a $2\times2$ maxpooling operation at the end of each block. The first block has $64$ filters in each layer, and this number is doubled after each block. The last feature map is vectorized by Global Average Pooling, which results in an embedding of $512$ dimensions. Image crops from bounding box annotations are rescaled to~$84\times84\times3$. 

Each model is trained for $30,000$ episodes with Adam optimizer initialized with a learning rate of $10^{-3}$ and is divided by a factor of $10$ every $10,000$ episodes. The validation set is used every $3000$ episodes to evaluate the mean accuracy of $500$ random episodes and early stopping is done based on the best validation accuracy.

\begin{table}[t]
\caption{\small Average Top-5 accuracy (\%) for $50$-way and $100$-way classification on Visual Genome (VG) and for $50$-way and $57$-way ($^\dagger$ all \textit{novel} classes) on Open Images (OI).}
    \centering
        \begin{tabular}{c|c|cccc}
        \hline
        & &\multicolumn{2}{c}{$50$-way}& \multicolumn{2}{c}{$100$-way/$57$-way$^\dagger$} \\
       \textbf{D}&\textbf{Model} & $1$-shot & $5$-shot & $1$-shot & $5$-shot\\
         \hline
      \multirow{3}{*}{VG} & PN \cite{snell2017prototypical} & $39.68$ & $59.52$ & $27.98$ & $47.23$\\
       &  AM3 \cite{xing2019adaptive} & $51.78$ & $64.14$ & $38.22$ & $51.57$ \\
    &     Ours & $\textbf{64.10}$ & $\textbf{72.86}$  & $\textbf{50.49}$ & $\textbf{61.19}$\\
         \hline
         \hline
      %   & &\multicolumn{2}{c}{} &  \multicolumn{2}{c}{}  \\
      % &\textbf{Model} & $1$-shot & $5$-shot &  $1$-shot & $5$-shot\\
      % \hline
      \multirow{3}{*}{OI} & PN \cite{snell2017prototypical}&$52.60$&$73.16$&$49.15$ &$70.69$\\
       &  AM3 \cite{xing2019adaptive}&$59.94$&$73.03$&$56.10$&$69.91$\\
    &     Ours&$\textbf{69.93}$&$\textbf{86.11}$&$\textbf{68.99}$&$\textbf{84.08}$\\
    \hline
    \end{tabular}
    \label{fig:50_100way}
    
\end{table}

\section{Results}
\begin{figure*}
    \centering
    \begin{subfigure}{0.2\linewidth}
    \includegraphics[width=\linewidth]{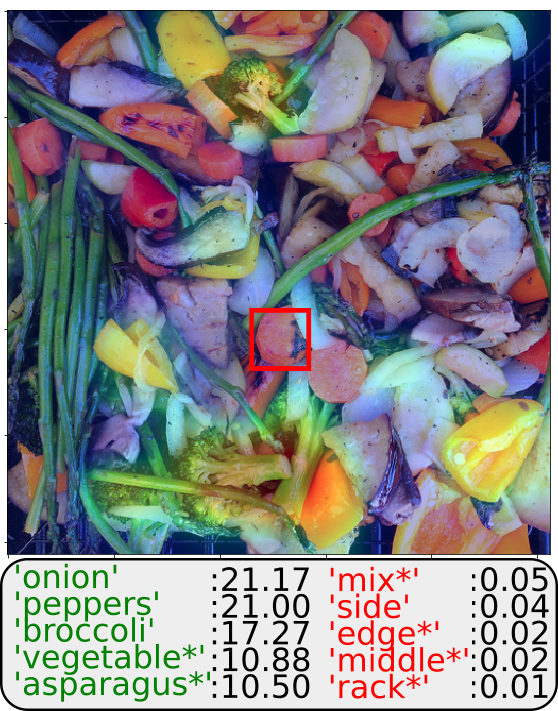}
    \caption{carrot}
    \label{ccama}
    \end{subfigure}
    \begin{subfigure}{0.2\linewidth}
    \includegraphics[width=\linewidth]{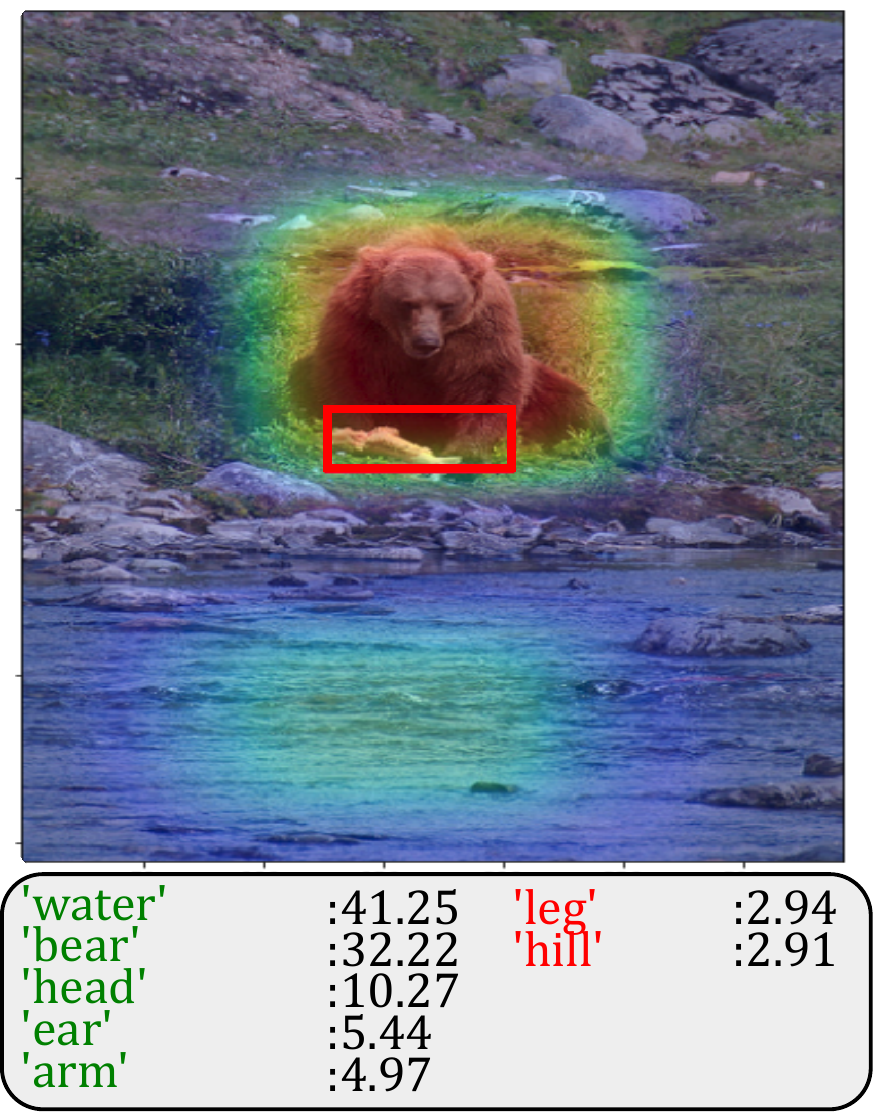}
    \caption{fish}
    \label{ccamb}
    \end{subfigure}
    \begin{subfigure}{0.2\linewidth}
    \includegraphics[width=\linewidth]{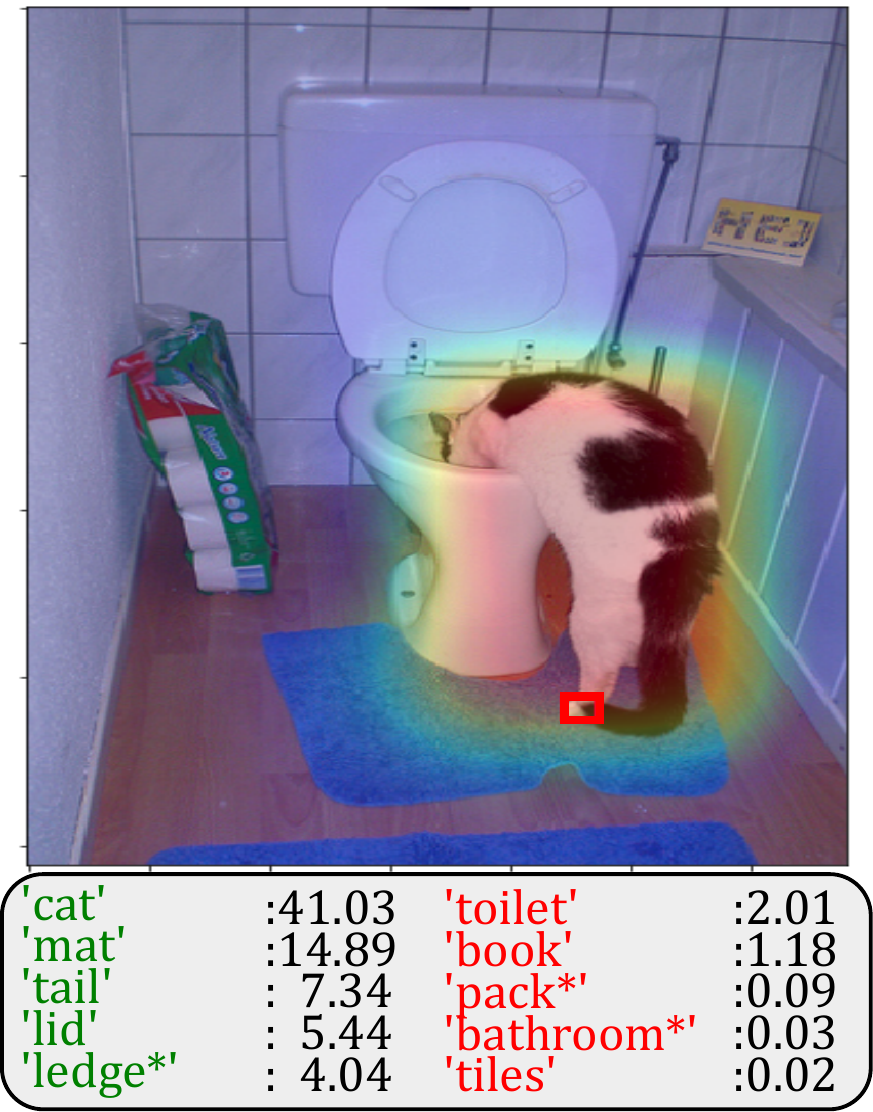}
    \caption{paw}
    \label{ccamc}
    \end{subfigure}
    \begin{subfigure}{0.2\linewidth}
    \includegraphics[width=\linewidth]{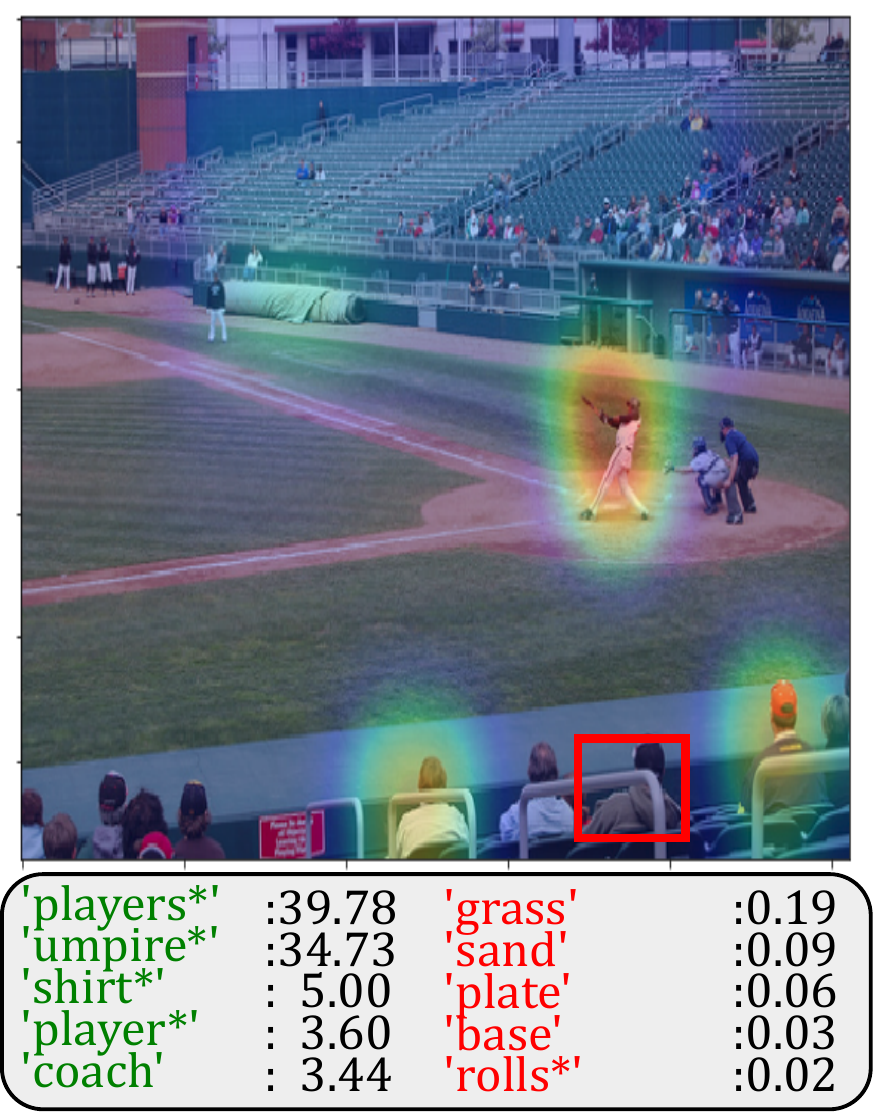}
    \caption{spectator}
    \label{ccamd}
    \end{subfigure}
    \caption{\small Illustration of the relative role of context estimated by CCAM. Five most important and least important co-occurrent concepts are shown in green and red, respectively, with associated weights (\%). Asterisks indicate that the word is a \textit{novel} class. Heatmaps are for visualization purposes. They are computed by summing the weight scores inside their respective bounding box annotations.}
    \label{fig:CCAM}
\end{figure*}
Since we want to study the contribution of using context information and class semantics in addition to the appearance of objects, we compare our approach to the following models that build on a Prototypical Networks (Protonet) backbone \cite{snell2017prototypical}. \textbf{ProtoNet} is our implementation of Prototypical Networks \cite{snell2017prototypical}. \textbf{AM3} is our implementation of the \textit{Adaptive Modality Mixture Mechanism} \cite{xing2019adaptive}, which supplement the support prototypes with the word embeddings of their corresponding class.

Table \ref{fig:openimages_res} shows the results of few-shot episodes on Visual Genome and Open Images. We also show Top-5 accuracy for $50$-way and $100$-way classification on Visual Genome and $50$-way and $57$-way classification on Open Images in Table \ref{fig:50_100way}. 

We can observe that the use of context information outperforms the visual-only ProtoNet \cite{snell2017prototypical} and the multimodal AM3 \cite{xing2019adaptive} by large margins. Even in the $1$-shot setting, where there is a risk of overfitting a particular context since there is only one example of scene, our results on both datasets show that it is still promising to use the context. Interestingly, our results support the benefits of using word embeddings as Xing \textit{et al.} \cite{xing2019adaptive} did with AM3, especially in the $1$-shot setting. Indeed, using word embeddings increases the accuracy in $5$-way and $20$-way by $10.27\%$ and $8.65\%$ on Visual Genome, respectively, and by $5.27\%$ and $5.32\%$ on Open Images, respectively. Our use of context further improves these results by an additional $9.04\%$ and $11.77\%$ on Visual Genome, and by $7.55\%$ and $11.36\%$ on Open Images for $5$-way and $20$-way classification, respectively.

Our model also performs reasonably well on larger-scale experiments shown in Table \ref{fig:50_100way}. With only one example per class in $100$-way classification on Visual Genome, our model almost doubles the Top-5 accuracy of ProtoNet \cite{snell2017prototypical}, with $50.49\%$ compared to $27.98\%$.

%\begin{table}[t]
%\caption{Top-5 accuracy for $100$-way classification}
%    \centering
%        \begin{tabular}{ccc}
%        \hline
%        \textbf{Model} & $1$-shot & $5$-shot \\
%         \hline
%         ProtoNet \cite{snell2017prototypical} & 27.98 & 47.23 \\
%         AM3-Proto \cite{xing2019adaptive} & 38.22 & 51.57 \\
%         Ours ($C_S$) & 50.49 & 61.19  \\
%         Ours ($C_T$) & 48.08 & 56.61  \\
%         Ours ($C_S \cup C_T$) & 52.20 & 60.25  \\
%         \hline
%    \end{tabular}
%    \label{fig:100way}
%\end{table}

\begin{table}[]
\caption{\small Average Top-1 accuracy (\%) during an ablation study for $5$-way classification on Visual Genome. V: Uses visual information; W: Uses word embeddings; C: Uses contextual information.}
    \centering
    \begin{tabular}{cccc||c|c||c|c}
    \hline
     & V & W & C & $C_{avg}$ & CCAM & $1$-shot & $5$-shot \\
      \hline 
     \small\textcolor{mygray}{\textit{1}}& \xmark & & & & &$52.23$&$69.37$\\
     \small\textcolor{mygray}{\textit{2}}& \xmark &\xmark & & & &$62.50$&$72.07$\\
     \small\textcolor{mygray}{\textit{3}}& \xmark & &\xmark &\xmark & &$61.20$ &$76.66$\\
     \small\textcolor{mygray}{\textit{4}}& \xmark &\xmark &\xmark &\xmark & &$69.10$&$76.63$\\
     \small\textcolor{mygray}{\textit{5}}& \xmark & &\xmark & &\xmark &$63.56$&$77.16$\\
     \small\textcolor{mygray}{\textit{6}}& \xmark &\xmark &\xmark & &\xmark &$71.54$&$78.50$\\
      \hline
    \end{tabular}
    \label{tab:ablation}
\end{table}

\paragraph{Ablation study.} We then performed an ablation study to examine the contribution of each modality and their interactions, shown in Table \ref{tab:ablation}. Interestingly, we observe a synergy between the use of word embeddings and contextual information, which is stronger in the $1$-shot setting. For instance, lines 2 and 3 show that the use of word embeddings or context alone perform similarly in $1$-shot, with $62.50\%$ and $61.20\%$ respectively; but when word embeddings and context and taken together, the accuracy significantly increases to $69.10\%$ (line 4) and even further to $71.54\%$ when CCAM is used (line 6). 

The ablation study also supports the benefits of considering the \textit{relative role of context} \cite{dohn2018concept}, as weighting the context elements with CCAM obtains better results than simply considering all of them equally (compare lines 3-4 with lines 5-6). We now investigate this aspect qualitatively. 

\begin{figure*}[t]
    \centering
    \includegraphics[width=\linewidth]{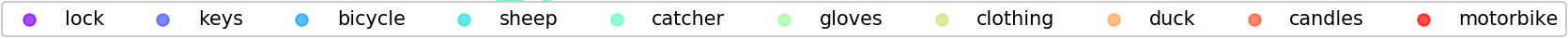}\\
    \begin{subfigure}{0.24\linewidth}
    \includegraphics[width=\linewidth]{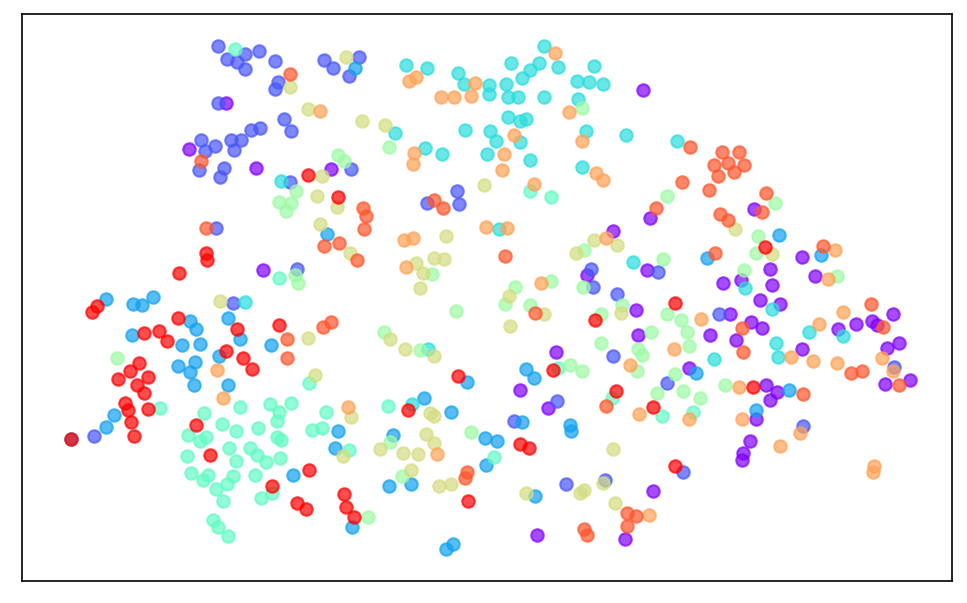}
    \caption{Image embeddings}
    \label{tsnea}
    \end{subfigure}
    \begin{subfigure}{0.24\linewidth}
    \includegraphics[width=\linewidth]{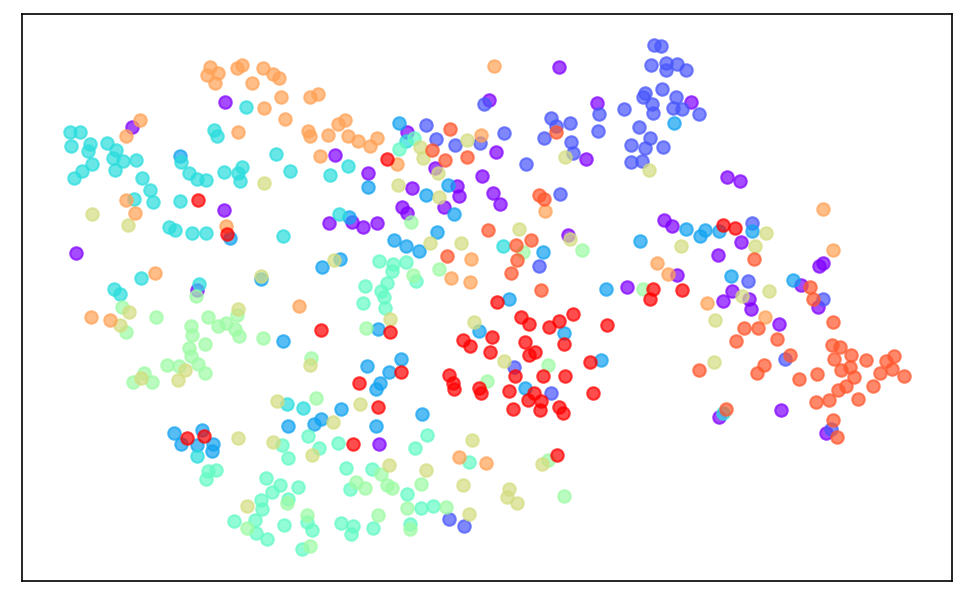}
    \caption{Context averaging}
    \label{tsneb}
    \end{subfigure}
    \begin{subfigure}{0.24\linewidth}
    \includegraphics[width=\linewidth]{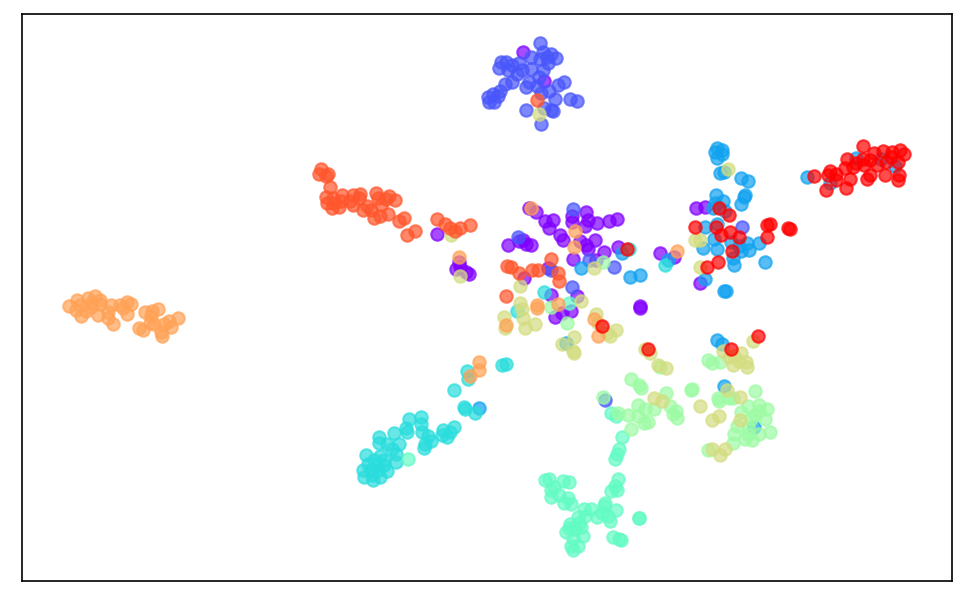}
    \caption{CCAM}
    \label{tsnec}
    \end{subfigure}
    \begin{subfigure}{0.24\linewidth}
    \includegraphics[width=\linewidth]{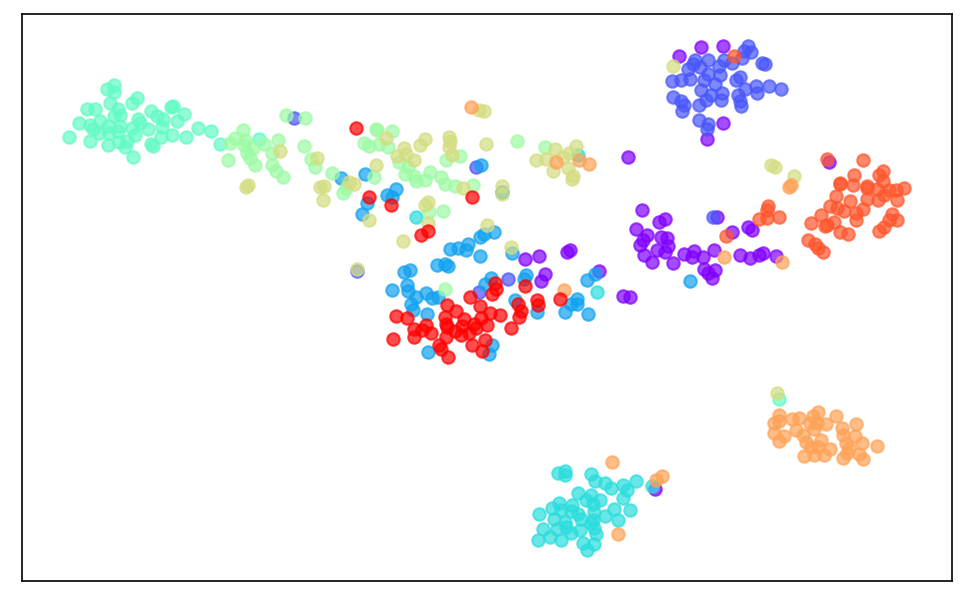}
    \caption{Context-Aware embeddings}
    \label{tsned}
    \end{subfigure}
    \caption{t-SNE visualization of different embeddings.}
    \label{fig:tsne}
\end{figure*}
\paragraph{Relative role of context \cite{dohn2018concept}.}
The relative role of context states that the importance of context elements is function of the focal object, which is supported by our results (see above). Fig. \ref{fig:CCAM} shows examples of CCAM outputs when our model needed to learn the concepts ``carrot'', ``fish'', ``paw'' and  ``spectator'', respectively. To further examine the ability of CCAM to meta-learn the importance of co-occurring concepts based on semantic similarity, we also included \textit{novel} classes in the formation of the context for these examples. We can see that CCAM correctly gives more weight to semantically relevant co-occurring concepts, even those that were never encountered during training (e.g. \textit{vegetable}, \textit{players} or \textit{umpire} that are \textit{novel} classes). On the other hand, CCAM also ignores background elements such as ``tiles'' in Fig. \ref{ccamc} and ``grass'' in Fig. \ref{ccamd}, as they are unlikely to help recognizing new instances of \textit{paw} and \textit{spectator} in query images.

To further study the contribution of context and CCAM, we show in Fig. \ref{fig:tsne} a t-SNE visualization \cite{tsne} of embeddings produced by our model using different amount of information. Visual embeddings (Fig. \ref{tsnea}) seem to produce ambiguous clusters, similar to context averaging (Fig. \ref{tsneb}) which considers each contextual item equally. On the opposite, our CCAM produces good clusters (Fig. \ref{tsnec}), which shows its ability, and the importance, to attend to discriminative elements. Also, some visually different objects seem to share similar contexts, as shown by the mixed cluster in the center of Fig. \ref{tsnec}. This problem is mostly solved by our GVSU that combines visual and contextual information (Fig. \ref{tsned}).

\begin{figure}
    \centering
    \includegraphics[width=0.85\linewidth]{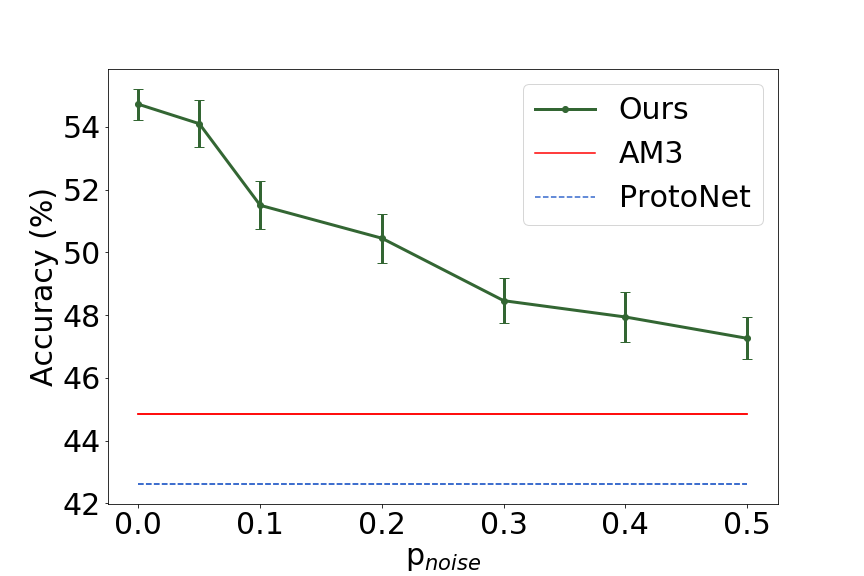}
    \caption{\small Robustness to noise in $20$-way $5$-shot tasks. Context elements in support and query sets are randomly swapped by another \textit{base} class label with a probability $p_{noise}$.}
    \label{fig:error}
\end{figure}

%\textcolor{red}{TODO : robustness to noise for $C_{avg}$}
\paragraph{Robustness to noise.} Since we aimed to study the potential contribution of context in a low-data regime, we used an oracle to annotate the context elements. In real applications, one would need to employ an object detection module to classify \textit{base} classes in query images and ideally in the few support examples also, although these could be manually annotated. While this aspect is left for future work, we evaluated the robustness of our model while adding noise in support and query context annotations to simulate an object detection module that makes a certain percentage of errors. With a probability $p_{noise}$, we randomly swap context elements by another \textit{base} class. The results for \mbox{$20$-way $5$-shot} classification with noise are shown in Fig. \ref{fig:error}. Protonet and AM3 are showed as reference but they are not influenced by noise in context annotations since they only use visual and visual+semantic information, respectively. Interestingly, our model seems reasonably tolerant to out-of-context objects. With $p_{noise} = 0.5$, our model still outperforms AM3~\cite{xing2019adaptive} and ProtoNet \cite{finn2018probabilistic}.
 
%\paragraph{Robustness to noise}

\paragraph{Learning semantic word embeddings through visual scenes.} Our approach also offers an auxiliary result that could be further investigated: CCAM implicitly learns and enriches semantic word embeddings by acting similarly to a CBOW model. To evaluate this aspect, we inputted to CCAM a matrix $S$ that contains all \textit{base} classes and we conditioned its attention on a few words $w$ to see how CCAM would weight~$S$. We show in Table \ref{tab:wordembeddings} a few examples. Interestingly, the contextual concepts defined by CCAM strongly differ from those obtained with Word2vec~\cite{mikolov2013distributed} embeddings, which shows that CCAM captures different semantic relations between concepts.

\section{Conclusion and Future work}
In this work, we proposed a few-shot learning model that uses scene context semantics to improve class representations. Our approach integrates the supplementary role of context \cite{dohn2018concept} by building context-aware class prototypes, and we apply the relative role of context \cite{dohn2018concept} with our CCAM, a module that proved to be able to focus on disciminative elements in the scene with respect to the class semantics (see Fig. \ref{fig:CCAM}, Fig. \ref{tsnec} and Table \ref{tab:wordembeddings}).

\begin{table}[t]
\caption{\small Examples of words and concepts that received the highest score by CCAM. Underlined words are also in the Top-10 of Word2vec \cite{mikolov2013distributed} cosine similarities.}
    \centering
    \begin{tabular}{|c|p{6cm}|}
    \hline
    \textbf{Word} & \textbf{Contextual words} \\
    \hline
\multirow{3}{*}{bike} & \small \underline{cyclists}, \underline{skateboards}, snowboards, guardrail, pedestrians, mopeds, snowmobile, tricycle, kiteboard, \underline{motorcycles} \\

 \hline
\multirow{2}{*}{rocks} & \small treetops, dunes, thickets, grassy, vegetation, hill, grasslands, sky, \underline{cliff}, pines \\
 \hline
\multirow{2}{*}{game} & \small referee, softball, scoreboard, basketball, jersey, volleyball, \underline{matches}, baseball, frisbee, football \\
 \hline
 
 \multirow{2}{*}{rope} &  \small boater, surfing, lifeguard, spear, \underline{ladders}, fisherman, horseback, \underline{ropes}, cliff, sandals \\
 \hline
 \multirow{2}{*}{sandwich} &  \small plate, plates, \underline{breads}, dough, \underline{sandwiches}, flatbread, cutlery, pork, pancakes, \underline{steak}, 
\\
 \hline
    \end{tabular}
    \label{tab:wordembeddings}
\end{table}

Our experiments on Visual Genome and Open Images showed promising results of using context information, by increasing the accuracy of Prototypical Networks \cite{snell2017prototypical} and AM3 \cite{xing2019adaptive} by large margins. More generally, our multi-semantics model is a step towards holistic approaches of few-shot object classification that can be applied in challenging real scenarios. 

As future work, we plan to replace ground-truth class annotations of context by automatic object detection and classification of base classes in queries.
Our experiment on robustness to noise suggests that even with many labeling errors, our method could still outperform models that ignore the context. We also plan to explore the ability of our model to \textit{learn semantic word embeddings through visual scenes}. This could be further investigated in line with work on learning multimodal word embeddings~\cite{lu2019vilbert, luddecke2019distributional, tan2019lxmert}.

{\small
\bibliographystyle{ieee_fullname}
\bibliography{egpaper.bbl}

\begin{thebibliography}{10}\itemsep=-1pt

\bibitem{al2016recovering}
Ziad Al-Halah, Makarand Tapaswi, and Rainer Stiefelhagen.
\newblock Recovering the missing link: Predicting class-attribute associations
  for unsupervised zero-shot learning.
\newblock In {\em Proceedings of the IEEE Conference on Computer Vision and
  Pattern Recognition}, pages 5975--5984, 2016.

\bibitem{arevalo2017gated}
John Arevalo, Thamar Solorio, Manuel Montes-y G{\'o}mez, and Fabio~A
  Gonz{\'a}lez.
\newblock Gated multimodal units for information fusion.
\newblock {\em arXiv preprint arXiv:1702.01992}, 2017.

\bibitem{bar2004visual}
Moshe Bar.
\newblock Visual objects in context.
\newblock {\em Nature Reviews Neuroscience}, 5(8):617, 2004.

\bibitem{Bateni_2020_CVPR}
Peyman Bateni, Raghav Goyal, Vaden Masrani, Frank Wood, and Leonid Sigal.
\newblock Improved few-shot visual classification.
\newblock In {\em IEEE/CVF Conference on Computer Vision and Pattern
  Recognition (CVPR)}, June 2020.

\bibitem{chen2017spatial}
Xinlei Chen and Abhinav Gupta.
\newblock Spatial memory for context reasoning in object detection.
\newblock In {\em Proceedings of the IEEE International Conference on Computer
  Vision}, pages 4086--4096, 2017.

\bibitem{chen2019image}
Zitian Chen, Yanwei Fu, Yu-Xiong Wang, Lin Ma, Wei Liu, and Martial Hebert.
\newblock Image deformation meta-networks for one-shot learning.
\newblock In {\em Proceedings of the IEEE Conference on Computer Vision and
  Pattern Recognition}, pages 8680--8689, 2019.

\bibitem{dohn2018concept}
Nina Dohn, Stig Hansen, and S{\o}ren Klausen.
\newblock On the concept of context.
\newblock {\em Education Sciences}, 8(3):111, 2018.

\bibitem{finn2017maml}
Chelsea Finn, Pieter Abbeel, and Sergey Levine.
\newblock Model-agnostic meta-learning for fast adaptation of deep networks.
\newblock In {\em Proceedings of the 34th International Conference on Machine
  Learning-Volume 70}, pages 1126--1135. JMLR. org, 2017.

\bibitem{finn2018probabilistic}
Chelsea Finn, Kelvin Xu, and Sergey Levine.
\newblock Probabilistic model-agnostic meta-learning.
\newblock In {\em Advances in Neural Information Processing Systems}, pages
  9516--9527, 2018.

\bibitem{gidaris2018dynamic}
Spyros Gidaris and Nikos Komodakis.
\newblock Dynamic few-shot visual learning without forgetting.
\newblock In {\em Proceedings of the IEEE Conference on Computer Vision and
  Pattern Recognition}, pages 4367--4375, 2018.

\bibitem{gupta2017distributed}
Abhijeet Gupta, Gemma Boleda, and Sebastian Pad{\'o}.
\newblock Distributed prediction of relations for entities: The easy, the
  difficult, and the impossible.
\newblock In {\em Proceedings of the 6th Joint Conference on Lexical and
  Computational Semantics (* SEM 2017)}, pages 104--109, 2017.

\bibitem{harris1954distributional}
Zellig~S Harris.
\newblock Distributional structure.
\newblock {\em Word}, 10(2-3):146--162, 1954.

\bibitem{jamal2019taml}
Muhammad~Abdullah Jamal and Guo-Jun Qi.
\newblock Task agnostic meta-learning for few-shot learning.
\newblock In {\em Proceedings of the IEEE Conference on Computer Vision and
  Pattern Recognition}, pages 11719--11727, 2019.

\bibitem{Jang_2020_WACV}
Ho-Deok Jang, Sanghyun Woo, Philipp Benz, Jinsun Park, and In~So Kweon.
\newblock Propose-and-attend single shot detector.
\newblock In {\em Proceedings of the IEEE/CVF Winter Conference on Applications
  of Computer Vision (WACV)}, March 2020.

\bibitem{johnson2002contextual}
Elaine~B Johnson.
\newblock {\em Contextual teaching and learning: What it is and why it's here
  to stay}.
\newblock Corwin Press, 2002.

\bibitem{krishnavisualgenome}
Ranjay Krishna, Yuke Zhu, Oliver Groth, Justin Johnson, Kenji Hata, Joshua
  Kravitz, Stephanie Chen, Yannis Kalantidis, Li-Jia Li, David~A Shamma,
  Michael Bernstein, and Li Fei-Fei.
\newblock Visual genome: Connecting language and vision using crowdsourced
  dense image annotations.
\newblock 2016.

\bibitem{OpenImages}
Alina Kuznetsova, Hassan Rom, Neil Alldrin, Jasper Uijlings, Ivan Krasin, Jordi
  Pont-Tuset, Shahab Kamali, Stefan Popov, Matteo Malloci, Alexander
  Kolesnikov, Tom Duerig, and Vittorio Ferrari.
\newblock The open images dataset v4: Unified image classification, object
  detection, and visual relationship detection at scale.
\newblock {\em IJCV}, 2020.

\bibitem{lake2015human}
Brenden~M Lake, Ruslan Salakhutdinov, and Joshua~B Tenenbaum.
\newblock Human-level concept learning through probabilistic program induction.
\newblock {\em Science}, 350(6266):1332--1338, 2015.

\bibitem{li2019large}
Aoxue Li, Tiange Luo, Zhiwu Lu, Tao Xiang, and Liwei Wang.
\newblock Large-scale few-shot learning: Knowledge transfer with class
  hierarchy.
\newblock In {\em Proceedings of the IEEE Conference on Computer Vision and
  Pattern Recognition}, pages 7212--7220, 2019.

\bibitem{li2019revisiting}
Wenbin Li, Lei Wang, Jinglin Xu, Jing Huo, Yang Gao, and Jiebo Luo.
\newblock Revisiting local descriptor based image-to-class measure for few-shot
  learning.
\newblock In {\em Proceedings of the IEEE Conference on Computer Vision and
  Pattern Recognition}, pages 7260--7268, 2019.

\bibitem{Li_2020_CVPR}
Yazhao Li, Yanwei Pang, Jianbing Shen, Jiale Cao, and Ling Shao.
\newblock Netnet: Neighbor erasing and transferring network for better single
  shot object detection.
\newblock In {\em IEEE/CVF Conference on Computer Vision and Pattern
  Recognition (CVPR)}, June 2020.

\bibitem{liu2018structure}
Yong Liu, Ruiping Wang, Shiguang Shan, and Xilin Chen.
\newblock Structure inference net: Object detection using scene-level context
  and instance-level relationships.
\newblock In {\em Proceedings of the IEEE conference on computer vision and
  pattern recognition}, pages 6985--6994, 2018.

\bibitem{lu2019vilbert}
Jiasen Lu, Dhruv Batra, Devi Parikh, and Stefan Lee.
\newblock Vilbert: Pretraining task-agnostic visiolinguistic representations
  for vision-and-language tasks.
\newblock In {\em Advances in Neural Information Processing Systems}, pages
  13--23, 2019.

\bibitem{luddecke2019distributional}
Timo L{\"u}ddecke, Alejandro Agostini, Michael Fauth, Minija Tamosiunaite, and
  Florentin W{\"o}rg{\"o}tter.
\newblock Distributional semantics of objects in visual scenes in comparison to
  text.
\newblock {\em Artificial Intelligence}, 274:44--65, 2019.

\bibitem{mikolov2013distributed}
Tomas Mikolov, Ilya Sutskever, Kai Chen, Greg~S Corrado, and Jeff Dean.
\newblock Distributed representations of words and phrases and their
  compositionality.
\newblock In {\em Advances in neural information processing systems}, pages
  3111--3119, 2013.

\bibitem{mottaghi2014role}
Roozbeh Mottaghi, Xianjie Chen, Xiaobai Liu, Nam-Gyu Cho, Seong-Whan Lee, Sanja
  Fidler, Raquel Urtasun, and Alan Yuille.
\newblock The role of context for object detection and semantic segmentation in
  the wild.
\newblock In {\em Proceedings of the IEEE Conference on Computer Vision and
  Pattern Recognition}, pages 891--898, 2014.

\bibitem{oliva2007role}
Aude Oliva and Antonio Torralba.
\newblock The role of context in object recognition.
\newblock {\em Trends in cognitive sciences}, 11(12):520--527, 2007.

\bibitem{oreshkin2018tadam}
Boris Oreshkin, Pau~Rodr{\'\i}guez L{\'o}pez, and Alexandre Lacoste.
\newblock Tadam: Task dependent adaptive metric for improved few-shot learning.
\newblock In {\em Advances in Neural Information Processing Systems}, pages
  721--731, 2018.

\bibitem{part2017incremental}
Jose~L Part and Oliver Lemon.
\newblock Incremental online learning of objects for robots operating in real
  environments.
\newblock In {\em 2017 Joint IEEE International Conference on Development and
  Learning and Epigenetic Robotics (ICDL-EpiRob)}, pages 304--310. IEEE, 2017.

\bibitem{Perez_Rua_2020_CVPR}
Juan-Manuel Perez-Rua, Xiatian Zhu, Timothy~M. Hospedales, and Tao Xiang.
\newblock Incremental few-shot object detection.
\newblock In {\em IEEE/CVF Conference on Computer Vision and Pattern
  Recognition (CVPR)}, June 2020.

\bibitem{rosenfeld2018elephant}
Amir Rosenfeld, Richard Zemel, and John~K Tsotsos.
\newblock The elephant in the room.
\newblock {\em arXiv preprint arXiv:1808.03305}, 2018.

\bibitem{schwartz2019baby}
Eli Schwartz, Leonid Karlinsky, Rogerio Feris, Raja Giryes, and Alex~M
  Bronstein.
\newblock Baby steps towards few-shot learning with multiple semantics.
\newblock {\em arXiv preprint arXiv:1906.01905}, 2019.

\bibitem{snell2017prototypical}
Jake Snell, Kevin Swersky, and Richard Zemel.
\newblock Prototypical networks for few-shot learning.
\newblock In {\em Advances in Neural Information Processing Systems}, pages
  4077--4087, 2017.

\bibitem{sung2018learning}
Flood Sung, Yongxin Yang, Li Zhang, Tao Xiang, Philip~HS Torr, and Timothy~M
  Hospedales.
\newblock Learning to compare: Relation network for few-shot learning.
\newblock In {\em Proceedings of the IEEE Conference on Computer Vision and
  Pattern Recognition}, pages 1199--1208, 2018.

\bibitem{tan2019lxmert}
Hao Tan and Mohit Bansal.
\newblock Lxmert: Learning cross-modality encoder representations from
  transformers.
\newblock In {\em Proceedings of the 2019 Conference on Empirical Methods in
  Natural Language Processing and the 9th International Joint Conference on
  Natural Language Processing (EMNLP-IJCNLP)}, pages 5103--5114, 2019.

\bibitem{tsne}
Laurens van~der Maaten and Geoffrey Hinton.
\newblock Visualizing data using {t-SNE}.
\newblock {\em Journal of Machine Learning Research}, 9:2579--2605, 2008.

\bibitem{vaswani2017attention}
Ashish Vaswani, Noam Shazeer, Niki Parmar, Jakob Uszkoreit, Llion Jones,
  Aidan~N Gomez, {\L}ukasz Kaiser, and Illia Polosukhin.
\newblock Attention is all you need.
\newblock In {\em Advances in neural information processing systems}, pages
  5998--6008, 2017.

\bibitem{vinyals2016matching}
Oriol Vinyals, Charles Blundell, Timothy Lillicrap, Daan Wierstra, et~al.
\newblock Matching networks for one shot learning.
\newblock In {\em Advances in neural information processing systems}, pages
  3630--3638, 2016.

\bibitem{wang2018zero}
Xiaolong Wang, Yufei Ye, and Abhinav Gupta.
\newblock Zero-shot recognition via semantic embeddings and knowledge graphs.
\newblock In {\em Proceedings of the IEEE Conference on Computer Vision and
  Pattern Recognition}, pages 6857--6866, 2018.

\bibitem{wang2019tafe}
Xin Wang, Fisher Yu, Ruth Wang, Trevor Darrell, and Joseph~E Gonzalez.
\newblock Tafe-net: Task-aware feature embeddings for low shot learning.
\newblock In {\em Proceedings of the IEEE Conference on Computer Vision and
  Pattern Recognition}, pages 1831--1840, 2019.

\bibitem{welinder2010caltech}
Peter Welinder, Steve Branson, Takeshi Mita, Catherine Wah, Florian Schroff,
  Serge Belongie, and Pietro Perona.
\newblock Caltech-ucsd birds 200.
\newblock 2010.

\bibitem{woo2018linknet}
Sanghyun Woo, Dahun Kim, Donghyeon Cho, and In~So Kweon.
\newblock Linknet: Relational embedding for scene graph.
\newblock In {\em Advances in Neural Information Processing Systems}, pages
  560--570, 2018.

\bibitem{xing2019adaptive}
Chen Xing, Negar Rostamzadeh, Boris~N Oreshkin, and Pedro~O Pinheiro.
\newblock Adaptive cross-modal few-shot learning.
\newblock {\em arXiv preprint arXiv:1902.07104}, 2019.

\bibitem{zablocki2019context}
Eloi Zablocki, Patrick Bordes, Benjamin Piwowarski, Laure Soulier, and Patrick
  Gallinari.
\newblock {Context-Aware Zero-Shot Learning for Object Recognition}.
\newblock In {\em {Thirty-sixth International Conference on Machine Learning
  (ICML)}}, Long Beach, CA, United States, June 2019.

\bibitem{zhang2019few}
Hongguang Zhang, Jing Zhang, and Piotr Koniusz.
\newblock Few-shot learning via saliency-guided hallucination of samples.
\newblock In {\em Proceedings of the IEEE Conference on Computer Vision and
  Pattern Recognition}, pages 2770--2779, 2019.

\end{thebibliography}
}

\end{document}